\documentclass[11pt,a4paper]{article}
\usepackage[hyperref]{emnlp-ijcnlp-2019}
\usepackage{times}
\usepackage{latexsym}
\usepackage{diagbox}
\usepackage{times}
\usepackage{latexsym}
\usepackage{graphicx}
\usepackage{amsmath}
\usepackage{float}
\usepackage{booktabs}
\usepackage{url}
\usepackage{subcaption}
\usepackage{caption}
\usepackage[shortlabels]{enumitem}
\usepackage{xcolor}
\usepackage{url}

\usepackage{mathtools}
\usepackage{placeins}
\usepackage{hyperref}
\usepackage{graphicx}
\graphicspath{ {./images/} }

\aclfinalcopy % Uncomment this line for the final submission

\newcommand{\Dearly}{$D_{\mathrm{early}}$}
\newcommand{\Dlate}{$D_{\mathrm{late}}$}
\newcommand{\Dmedian}{$D_{\mathrm{med}}$}
\newcommand*\samethanks[1][\value{footnote}]{\footnotemark[#1]}
\DeclareMathOperator{\KL}{KL}

\DeclarePairedDelimiterX{\infdivx}[2]{(}{)}{%
  #1\;\delimsize\|\;#2%
}
\newcommand{\infdiv}{D_{\KL} \infdivx}

\title{Countering the Effects of Lead Bias in News Summarization via Multi-Stage Training and Auxiliary Losses}

\author{ Matt Grenander\thanks{\ \ Equal contribution.}, Yue Dong\samethanks \\
  McGill University / MILA \\
  \small \tt \{matthew.grenander,\\ \small \tt yue.dong2\}\small \tt@mail.mcgill.ca \\\And
  Jackie C. K. Cheung \\
  McGill University / MILA \\
  \small \tt jcheung@cs.mcgill.ca \\\And 
  Annie Louis \\
  University of Edinburgh \\
  Alan Turing Institute \\
  \small \tt alouis@inf.ed.ac.uk}

\date{}

\begin{document}
\maketitle
\begin{abstract}
Sentence position is a strong feature for news summarization, since the lead often (but not always) summarizes the key points of the article. In this paper, we show that recent neural systems excessively exploit this trend, which although powerful for many inputs, is also detrimental when summarizing documents where important content should be extracted from later parts of the article. We propose two techniques to make systems sensitive to the importance of content in different parts of the article. The first technique employs `unbiased' data; i.e., randomly shuffled sentences of the source document, to pretrain the model. The second technique uses an auxiliary ROUGE-based loss that encourages the model to distribute importance scores throughout a document by mimicking sentence-level ROUGE scores on the training data. We show that these techniques significantly improve the performance of a competitive reinforcement learning based extractive system, 
with the auxiliary loss being more powerful than pretraining. 
%For documents where the important content appears later in the article, the improvement is even better.
%AL2 commenting out the above since we are doing worse on dearly. The above would imply that we are improving on all types of documents
\end{abstract}

\section{Introduction}
\label{sec:intro}
%Automatic summarization, the task of shortening a text while preserving its main content, can be divided into two categories: extractive and abstractive. While abstractive summaries may potentially be more concise through the use of paraphrasing, extractive methods, which copy text snippets from the source document, are still very attractive. Compared to their abstractive competitors, extractive models are usually simpler, quicker, more grammatical and often produce more reliable outputs in terms of consistency with the source text.

Extractive summarization remains a simple 
and fast approach to produce summaries which 
are grammatical and accurately represent the source text. 
%AL2 are current neural extractive systems faster than abstractive?
% MG: I think hard to say, not all papers report total time to train...
%The central component of extractive summarization is content selection, usually accomplished through sentence extraction. 
In the news domain, these systems are able to 
use a dominant signal: the position of a sentence 
in the source document. 
%Sentence position is often a dominant signal for sentence selection in news corpora 
Due to journalistic conventions which place important information
early in the articles, the lead sentences often contain key information. In this paper, we explore how systems can look beyond this simple trend. 

Naturally, automatic
systems have all along exploited position cues in news 
as key indicators of important content  \citep{Schiffman:2002:EMS,hong2014improving,ext_bert}. The `lead' baseline is rather strong in single-document news summarization \citep{brandow1995automatic,nenkova2005automatic}, 
with automatic systems only modestly improving the 
results. 
%As a result, a `lead' baseline using the first few sentences of the article is often hard for automatic systems to beat 
Nevertheless, more than 20-30\% of %extractive gold training labels
summary-worthy sentences come from the second half of news documents \citep{data2_nallapati2016abstractive,kedzie2018content}, 
and the lead baseline, as shown in Table \ref{tab:lead_ex}, does not always produce convincing summaries.
%So to achieve good performance, in particular, going beyond the lead
%baseline, 
So, systems must balance the position bias with representations of the semantic content 
throughout the document. Alas, preliminary studies
\citep{kedzie2018content} suggest that even the most recent neural 
methods predominantly pick sentences from the lead, and 
that their content selection performance drops greatly
when the position cues are withheld. 
%AL2 can we compute percentage of sentences not from the lead
%after doing KL. It might be we are picking good sentences
%but content-wise they are same as lead, so we dont get
%improved ROUGE scores? MG: added as a column in the results table

\begin{table}[t]
    \centering
    \small
    \begin{tabular}{|p{0.94\linewidth}|}
        \hline
        \textbf{Lead-3:} Bangladesh beat fellow World Cup quarter-finalists Pakistan by 79 runs in the first one-day international in Dhaka. Tamim Iqbal and Mushfiqur Rahim scored centuries as Bangladesh made 329 for six and Pakistan could only muster 250 in reply. Pakistan will have the chance to level the three-match series on Sunday when the second odi takes place in Mirpur. \\ \hline
        \textbf{Reference:} Bangladesh beat fellow World Cup quarter-finalists Pakistan by 79 runs. Tamim Iqbal and Mushfiqur Rahim scored centuries for Bangladesh. Bangladesh made 329 for six and Pakistan could only muster 250 in reply. Pakistan will have the chance to level the three-match series on Sunday. \\ \hline \hline
        \textbf{Lead-3}: Standing up for what you believe. What does it cost you? What do you gain? \\ \hline
        \textbf{Reference:} Indiana town's Memories Pizza is shut down after online threat. Its owners say they'd refuse to cater a same-sex couple's wedding. \\
        \hline
    \end{tabular}
    \caption{`Lead' (first 3 sentences of source) can produce extremely faithful (top) to disastrously inaccurate (bottom) summaries. Gold standard summaries are also shown.}
    \label{tab:lead_ex}
\end{table}

In this paper, we verify that 
%the ability of recent extractive models to balance semantic content versus sentence position. Do 
%whether 
sentence position and lead bias dominate the learning signal for  state-of-the-art neural extractive summarizers in 
the news domain. 
%current models on news. 
%, and if so, is this dominant signal always appropriate? 
We then present techniques to improve content selection in 
the face of this bias. 
%n a better representation of semantic content going beyond %positional cues?
%To this end, we analyze state-of-the-art neural extractive summarizers in the news domain.
%, considering specifically the impact of sentence position. 
%By manipulating the input data, we verify that sentence position bias dominates the learning signal for extractive models to the point that models largely ignore the articles' content. 
%We propose two methods to counter the effects of this issue.
The first technique makes use of `unbiased data' 
created by permuting the order of sentences in the training
articles. We use this shuffled dataset for pre-training, followed by 
training on the original (unshuffled) articles. The second method 
introduces an auxiliary loss which encourages the model's scores
for sentences to mimic an estimated score distribution over the
sentences, the latter computed using ROUGE overlap with the gold standard. 
We implement these techniques for two recent reinforcement learning based systems, RNES \citep{DBLP:conf/aaai/WuH18} and BanditSum \citep{dong2018banditsum}, and evaluate them on the CNN/Daily Mail dataset \citep{hermann2015teaching}.
 We find that our
auxiliary loss achieves significantly better ROUGE scores
compared to the base systems, and that the improvement is even 
more pronounced when the true 
best sentences appear later in the 
article. On the other hand, the pretraining approach
produces mixed results.  
%AL2 have we actually measured time?
We also confirm that when summary-worthy sentences appear late, there is a large performance discrepancy between the oracle summary and state-of-the-art summarizers,
%the best scores achievable are too high, 
indicating that learning to balance lead bias with other features of news text is a noteworthy issue to tackle.
% this problem of position bias is a noteworthy issue to tackle.

%AL2 to update once we have final results

%Our methods for countering these effects achieve statistically significant improvement on a benchmark summarization dataset, with even greater performance gains when the best extractive indices appear later in the article.

% Our main results reveal the following: 1) Sentence position is a dominant feature in current extractive summarization models.
% 2) Current systems do not learn an appropriate amount of lead bias. This phenomenon hurts the performance of models in the cases where good summary sentences appear later.
% 3) Training models with randomly shuffled data (within the document) in a multi-task learning setting or adding an ROUGE-based entropy regularizer encourages the model to make use of the actual semantic content of the article, and the performance improvement is statistically significant on CNN/Daily Mail.

\section{Related Work}\label{sec:related_work}
%Classical summarization methods \citep[e.g., ][]{tra_ext1_luhn1958automatic,tra_ext2_gong2001generic, tra_ext4_conroy2001text,tra_ext5_wong2008extractive} have largely been tested on small news corpora, such as the DUC and TAC datasets (\textcolor{blue}{cite these}). More recent summarization datasets likewise fall into the news domain \cite{sandhaus2008new, grusky2018newsroom, narayan2018don}. The most widely-used benchmark dataset is the CNN / Daily Mail set, introduced by \citet{hermann2015teaching}. Since its release,

Modern summarization methods for news are typically
based on neural network-based sequence-to-sequence learning
\cite{cnn1_kalchbrenner2014convolutional,cnn2_kim2014convolutional,rnn2_chung2014gru,ext2_2015Yin,ext3_cao2015learning,ext4_cheng2016neural,ext5_summarunner,narayan2018don,neusum}.
%neural networks, and typically employ some form of encoder-decoder structure,
%have been proposed, 
% with either convolution neural networks (CNN) \cite{cnn1_kalchbrenner2014convolutional,cnn2_kim2014convolutional,ext2_2015Yin,ext3_cao2015learning,narayan2018don}, recurrent neural networks \cite{rnn2_chung2014gru,ext4_cheng2016neural,ext5_summarunner,dong2018banditsum, neusum} or a combination of the two \cite{DBLP:Narayan/2018,DBLP:conf/aaai/WuH18}.
In MLE-based training, extractive summarizers are trained with gradient ascent to maximize the likelihood of heuristically-generated ground-truth binary labels \citep{ext5_summarunner}. Many MLE-based models do not perform as well as their reinforcement learning-based (RL) competitors that directly optimize ROUGE \cite{abs5_paulus2017deep,DBLP:Narayan/2018,dong2018banditsum,DBLP:conf/aaai/WuH18}. 
%The lower performance of MLE may be due to exposure bias \citep{Ranzato2015SequenceLT}, and the inaccuracy of the created extractive labels \citep{arumae2018reinforced}. 
As RL-based models represent the state of the art for extractive summarization, we analyze them in this paper.

The closest work to ours is a recent study by \citet{kedzie2018content} which showed that 
MLE-based models
learn a significant bias for selecting early sentences
when trained on news articles as opposed  
to other domains. As much as 58\% of selected
summary sentences come directly from the lead. 
Moreover, when these models
are trained on articles whose sentences
are randomly shuffled, the performance drops considerably for news domain only. 
%In this work, we corroborate \citet{kedzie2018content}'s 
%finding via thorough experiments. 
While this drop 
%drop observed by \citet{kedzie2018content} 
%can arise from the destruction 
%of sentence position cues, it may also come from degeneration %of the article's coherence and context. 
could be due to the destruction of position cues, 
it may also arise because the article's coherence
and context were lost. 

In this paper, we employ finer control on the distortion 
of sentence position, coherence, and context, and confirm that performance 
drops are mainly due to the lack of position cues. 
We also propose the first techniques to 
%further extend this by proposing effective methods to 
counter the effects of lead bias in neural extractive systems.

% Recent work by \citet{kedzie2018content} trained models on randomly shuffled news data and observed performance drops. This performance drop can result from the destruction of sentence position cues, but can also come from the degeneration of the coherence and the context. We instead design a series of experiments that controls the level of distortions on sentence positions, coherence, and contexts. Our empirical results  verified that the performance drop mainly results from the  destruction of lead bias.  In addition, we focus on how to counter the lead bias effects in extractive summarizers and propose simple but effective methods based on random shuffled data and entropy regularization (e.g. \citet{a3c, entropy_reg}).

%Entropy regularization has long been used in the context of both reinforcement learning and supervised learning to prevent overconfident models (e.g. \citet{a3c, entorpy_reg}). Our auxiliary loss method loosely resembles using an entropy regularizer in purpose; however, our proposed method gives a much more directed signal to the learning agent than simply encouraging higher entropy policies. To back our claim, we show that our method surpasses a simple entropy regularization technique.  
%MG: I'm running this now after Yue/Jackie suggested it so we'll see about the above claim
%Yue: this should go somewhere other than  related work?

\section{Base Models for Extractive Summarization}

%In extractive summarization, the goal is to select text snippets---typically sentences---to form a summary. 

In supervised systems, given a document $D = \{ s_1, . . . , s_n \}$ with $n$ sentences, 
a summary can be seen as
set of binary labels $y_1,\dots, y_n \in \{0, 1\}$, where $y_i = 1$ 
indicates that the $i$-th sentence is included in the summary.

We choose to experiment with two state-of-the-art RL-based extractive models: 
{\bf RNES} \cite{DBLP:conf/aaai/WuH18} and {\bf BanditSum} \citep{dong2018banditsum}. Both employ an encoder-decoder structure, where the encoder extracts sentence features into fixed-dimensional vector representations $h_1, \dots,h_n$, and a decoder produces the labels $y_1,\dots, y_n$ based on these sentence representations. RNES uses a CNN+bi-GRU encoder, and
BanditSum a hierarchical bi-LSTM. RNES's decoder is 
\textit{auto-regressive}, meaning it predicts the current sentence's label based on decisions made on previous sentences; i.e., $y_t = f(D,h_t, y_{1:t-1})$. In BanditSum, there is no such dependence: it produces affinity
scores for each sentence and the top scoring sentences are then selected.

\section{Lead Bias of News Systems}\label{sec:lead_bias}

\begin{table*}[h]
\centering
\small
\begin{tabular}{c|ccccc|cc}
\toprule
\diagbox{train setting}{test setting} &original & random & reverse & insert-lead & insert-lead3 & Mean & Std. Dev.  \\ \hline
Lead-3 baseline &32.68 & 22.81&17.94&27.67&27.68 &25.76 &5.00\\ \hline
original & \textbf{33.85} & 26.18 &20.71 &31.71 & 31.11 &28.71 & 4.72\\ 
random & 30.88 & \textbf{29.70} & 29.79 & 29.97 & 30.09 &\textbf{30.09}& \textbf{0.42}\\ 
reverse & 21.35 & 26.32 & \textbf{33.59} & 21.63 & 21.65 &24.91 & 4.72 \\ 
insert-lead & 33.21 & 26.07 & 20.70 & \textbf{33.41} & 31.59 &29.00 &4.93 \\ 
insert-lead3 & 32.29 & 25.57 & 20.22 & 32.92 & \textbf{32.15} &28.63&4.98 \\ \bottomrule
%32.73
\end{tabular}

\caption{BanditSum's performance---calculated as the average between ROUGE-1,-2, and -L F1---on the validation set of the CNN/Daily Mail corpus. The sentence position information is perturbed at different levels, as explained in Section \ref{sec:lead_bias}.} %In random setting, sentences are shuffled randomly; in reverse mode, we reverse the sentences in the documents; in insert-lead and insert-lead3, we insert an out-of-document sentence as the lead sentence or randomly as one of the first three sentences, respectively. } The mean and variance report model trained on different data performing differently across all settings. Mean and variance are measured across different validation sets .
\label{tab:data_manipulation}
\end{table*}

First, we investigate the impact of sentence 
position on our models. We manipulate the \textbf{original}
CNN/Daily Mail dataset to 
preserve sentence position information
at different levels. In the \textbf{random} setting, sentences are shuffled randomly; in \textbf{reverse}, they are in reverse order; in \textbf{insert-lead} and \textbf{insert-lead3}, we insert an out-of-document 
sentence (chosen randomly from the corpus) as the first sentence or randomly as one of the first three sentences, respectively. %Finally, \textbf{original} preserves the original ordering.

In Table \ref{tab:data_manipulation}, we show BanditSum's performance,\footnote{We notice the same trends on  RNES.
% , though RNES drops 4.2 and Refresh drops 3.4 points in average ROUGE when trained on shuffled data and tested on the original dataset.
} when trained and tested on the various datasets. 
All models (except random)  perform worse when tested on a mismatched data perturbation. 
Even when the distortion is at a single lead 
position in \textbf{insert-lead} and \textbf{insert-lead3}, 
the performance on the original data is significantly
lower than when trained without the distortion.
%Interestingly, the model trained on randomly shuffled data performs consistently regardless of how the data is perturbed. 
These results corroborate \citet{kedzie2018content}'s findings for RL-based
systems. Interestingly, the \textbf{random} model has the 
best mean performance and the lowest variation indicating that completely removing
the position bias may allow a model to focus on learning robust sentence semantics.

\section{Learning to Counter Position Bias}

%, we focus on extracting sentences to form summaries. Thus, the basic text snippet unit is a sentence, and we impose three sentences as the budget.} %Each sentence consists $|s_i|$ words, which usually transformed to a sequence of word embeddings $s_i = w ^{(i)}_1, . . . , w^{(i)}_{|si|}$.  

%We usually impose a constraint of word budget $c \in \mathbb{N}$ on the length of the summary, which can be expressed as $\sum_i^n y_i \cdot |s_i| \leq c$.  
% We test state-of-the-art models that are auto-regressive  as a sequence tagging problem, following (Conroy and O’Leary, 2001). We also test non-auto-regressive models as they perform better or as good as auto-regressive models in several researches. 

% We tested three state-of-the-art extractive models, one auto-regressive model -- RNES \cite{DBLP:conf/aaai/WuH18} and two non-auto-regressive models -- BanditSum \citep{dong2018banditsum} and Refresh \citep{narayan2018don}. The following table specifies the encoder-decoder choice of each model. 

%  \vspace{-mm}

We present two methods which encourage models 
to locate key phrases at diverse
parts of the article.

\subsection{Multi-Stage Training}
This technique is inspired by the robust results from the
\textbf{random} model in section  \ref{sec:lead_bias}. 
% in the previous section 
% which indicate that a model trained on randomly shuffled sentences
% may be learning sentence semantics more generally compared to the
% original articles. However, inter-sentence coherence is affected
% by the shuffling.
We implement a
multi-stage training method for both BanditSum and RNES where 
in the first few epochs, we
%AL2 verify that pretraining is first two epochs, MG: it is!
train on an `unbiased' dataset where the sentences in every training document
are randomly shuffled. We then fine-tune
the models by training on the original training articles. 
% doing so, we preserve sentence-internal semantics while removing the possibility 
% of learning based on the sentence position. As the formulation 
% of ROUGE does not account for sentence ordering, randomly permuting sentences 
% does not affect the overall ROUGE score of the selected sentences. And although 
% inter-sentence coherence is lost while pre-training, it can still be learned 
% while training on \Dinorder{}. 
The goal is to prime the model to learn
sentence semantics independently of position, and then  
introduce the task of balancing semantics and positional cues. 

%Based on our experimental results, although the non-auto-regressive models seem to have less lead bias, they seem still suffer a significant loss when tested on the documents where sentence position clues are removed. This result suggests that models are still using more than the appropriate level of lead bias in their decision making. In this work, we propose two approaches to counter this bias. 

% Central to our first approach is the use of an `unbiased' dataset obtained by randomly shuffling the sentences within a document, \Drandom{} (c.f., the original dataset, \Dinorder{}). %We employ a scheduler \citep{kiperwasser2018scheduled} in a multi-task setting, where we feed sentences from \Drandom{} and \Dinorder{} in an interleaved manner, gradually feeding less data from \Drandom{} with probability $p = (1/2)^{epoch-1}$.
% We pre-train the  model  on \Drandom{} and then fine-tune on \Dinorder{}. By doing so, we preserve sentence-internal semantics while removing the possibility of learning based on the sentence position. (more details in Table \ref{tab:data_manipulation}). As the ROUGE evaluation does not take sentence order into account, randomly shuffling the sentences will not affect ROUGE scores of selected sentences. Although between-sentence coherence information is lost in \Drandom{}, it can still be learned and used while learning on \Dinorder. We hope to prime the model with \Drandom{} to pay attention to contents, in balancing to sentence position and discourse structure mostly learned on \Dinorder. 

\subsection{ROUGE-based Auxiliary Loss}
We observed that BanditSum tends to converge to a low-entropy policy, in the sense that the model's affinity scores are either 1 or 0 at the end of training.
Furthermore, over 68\% of its selections are from the three leading sentences of the source.
%AL2 can you find the above number for RNES?
%(\citet{kedzie2018content} report 58\% for a MLE based summarization system.)
%and that average affinity scores for BanditSum are far higher in the first three indices than for other sentences. 
%In RL studies,
Regularizing low-entropy policies can increase a model's propensity to explore potentially good states 
or stay close to a known good policy \citep{nachum2017improving,galashov2019information}. 
%AL2 i think the cited nachum2017bridging is the wrong paper. It should be 
% O. Nachum, M. Norouzi, and D. Schuurmans. Improving policy gradient by exploring under-
% appreciated rewards. ICLR, 2017. I've changed: make sure this is right. 
We extend this idea to summarization by introducing a ROUGE-based loss 
which regularizes the model policy using an estimate of the value of individual sentences.

% Another observation we made from the original BanditSum is that the model converges to a deterministic policy similar to lead, evidenced  by the prediction distribution on the validation set has a low normalized entropy (see Table \ref{tab:entropy} for a comparison). This is not ideal for summarization, where distinct documents should have a different distributions. Instead, letting an RL-based model converges to a stochastic policy with an entropy regularizer would improve the generalization performance and encourage model to explore \citep{nachum2017bridging, galashov2019information}. We further extend this idea to introduce a ROUGE-based loss that not only acts as an a regularizer, but also considers how much content each sentence carries.

These sentence-level estimates are computed as a \textit{distribution} $P_R$:

\begin{equation}
    P_R( x = i ) = \frac{r(s_i, \mathcal{G})}{\sum_{j=1}^{n}{r(s_j, \mathcal{G})}},
\end{equation}

\noindent where $r$ is the average of ROUGE-1, -2 and \mbox{-L} F\textsubscript{1} scores between sentence $s_i$ in the article
and the reference summary $\mathcal{G}$. We would like the model's predictive distribution $P_\mathcal{M}$ to approximately match
% the sentence-level ROUGE distribution 
$P_R$. To compute $P_\mathcal{M}$, 
%=q_1, \ldots, q_n$
we normalize the predicted scores from a non-auto-regressive model. In an auto-regressive model such as RNES,  the decision of including a sentence 
depends on those selected so far. So a straightforward KL objective is hard to implement, and 
we use this technique for BanditSum only.

Our auxiliary loss is defined as the KL divergence: $\mathcal{L}_{\KL} = \infdiv{P_R}{P_\mathcal{M}}$. The update rule then becomes:
%\vspace{-3mm}
\begin{equation}
\label{eq:kl_loss}
    \theta^{(t+1)} = \theta^{(t)} + \alpha \left( \nabla \mathcal{L}_{\mathcal{M}}(\theta^{(t)}) + \beta \nabla \mathcal{L}_{\KL}(\theta^{(t)}) \right)
\end{equation}

where $\theta^{(t)}$ represents the model's parameters at time step $t$, $\mathcal{L}_{\mathcal{M}}$ is the original model's loss function, and $\beta$ is a hyperparameter.

\section{Experimental Setup}
We use the CNN/Daily Mail dataset \citep{hermann2015teaching} with the standard train/dev/test splits of 287,227/13,368/11,490. 
%AL2 can we write overall numbers combining the CNN/Daily Mail or is this how these are typically reported?
% MG: I think both ways are ok, some papers don't even list the splits. I combined them.
To avoid inconsistencies, we built on top of the author-provided implementations for 
BanditSum and our faithful reimplementation of RNES. 

To reduce training time, we pre-compute and store the average of ROUGE-1, -2, and -L for every sentence triplet of each article, using a HDF5 table and PyTables \cite{pytables, hdf5}. This allows for a considerable increase in training speed. We limit the maximum number of sentences considered in an article to the first 100.

All the models were trained for 4 epochs. For the multi-stage training, 
%we try both an auto-regressive model - RNES \cite{DBLP:conf/aaai/WuH18} - and a non-auto-regressive model - BanditSum \cite{dong2018banditsum}. We 
we pretrain for 2 epochs, then train on the original articles for 2 epochs.
 We set the auxiliary loss hyperparameters $\alpha=1e-4$ and $\beta=0.0095$ in 
eq. \ref{eq:kl_loss} based on a grid search using the Tune library \cite{tune}. 
%We train models for 4 epochs after observing that BanditSum performance continues to improve beyond 2 epochs.

We also train a baseline {\bf entropy} model by replacing $\mathcal{L}_{\KL}$ with the negated entropy of $P_\mathcal{M}$ in eq. \ref{eq:kl_loss}.
%to confirm that our method is not acting as a simple entropy regularizer. 
This loss penalizes low entropy, helping the model explore, but it is `undirected' compared to our proposed method. We present the results of Lead-3 baseline (first 3 sentences), and two other competitive
models---Refresh\footnote{We are unable to evaluate this model on the lead overlap measure due to lack of access to the model outputs.} \citep{narayan2018don} and NeuSum \citep{neusum}.

Lastly, we include results from an \textit{oracle} summarizer, computed as the triplet of source sentences with the highest average of ROUGE-1, -2 and -L scores against the abstractive gold standard.

\section{Results and Discussion} \label{sec:result}

Table \ref{tab:results} reports the F1 scores for ROUGE-1,-2 and -L \cite{eva1_lin:2004:ACLsummarization}. 
We use the \emph{pyrouge}\footnote{\url{www.github.com/bheinzerling/pyrouge}} wrapper library to evaluate the final models, while training
with a faster Python-only 
implementation\footnote{\url{www.github.com/Diego999/py-rouge}}. %AL2 give link to python implementation. MG: done.

%In addition, \citet{kedzie2018content} has shown that the performance of auto-regressive model drops around 4 ROUGE average F1 point if they are trained on randomly shuffled data and tested on the unshuffled data. From Table \ref{tab:data_manipulation}, we noticed that this number only drops by 3 ROUGE points in non-auto-regressive models. Thus, it seems that non-auto-regressive models has less lead bias.
 
\begin{table}[t]
    \centering
    \small
    \begin{tabular}{l|ccc|c}
        \toprule						
        Model	&	\multicolumn{3}{c}{ROUGE}	& Overlp	\\
        	&	1	&	2	&	L	&	\%	\\ \hline
        Lead-3	&	40.06	&	17.53	&	36.18	&   100.0	\\
        Oracle	&	56.53	&	32.65	&	53.12	&	27.24	\\
        Refresh	&	40.0	&	18.2	&	36.6	&	--	\\
        NeuSum	&	40.15	&	17.80	&	36.63	&	58.24	\\\hline
        RNES	&	41.15	&	18.81	&	37.75	&	68.44	\\ 
        RNES+pretrain & 41.29    &  18.85         &    37.79       &     68.22   \\ \hline 
        BanditSum	    &	41.68	&	18.78	&	38.00	&	69.87	\\
        B.Sum+pretrain  &   41.68   &   18.79   &   37.99   &   70.77   \\ %AL3 4 epoch results
        B.Sum+entropy	&	41.71	&	18.87	&	38.04	&	64.83	\\
        BanditSum+KL	&	\textbf{41.81*}	&	\textbf{18.96*}	&	\textbf{38.16*}	&	65.13	\\
        \bottomrule
    \end{tabular}
    \caption{ROUGE scores for systems. `Overlp' denotes the model's overlap in extraction choices with the lead-3 baseline. Scores significantly higher than 
    BanditSum  with $p<0.001$ (bootstrap resampling test) are marked with *.}
    \label{tab:results}
\end{table}

% \begin{table}
%     \centering
%     \small
%     \begin{tabular}{l|ccc}
%         \toprule							
%         Model	&	\multicolumn{3}{c}{ROUGE}	\\				
%         	&	1	&	2	&	L	\\ \midrule
%         Lead3	&	40.06	&	17.53	&	36.18	\\
%         Refresh	&	40.0	&	18.2	&	36.6	\\
%         NeuSum	&	41.59	&	19.01	&	37.98	\\ \hline
%          \multicolumn{4}{c}{2 epochs}\\ \hline 
%         RNES	&	41.15	&	18.81	&	37.74	\\
%         RNES+pretrain	&	41.26	&	18.85	&	37.79	\\ 
%         BanditSum	&	41.5	&	18.7	&	37.6	\\
%         BanditSum+pre-train    & 41.64   & 18.76  & 37.81 \\ 
%         \hline
%          \multicolumn{4}{c}{4 epochs}\\ \hline
%         BanditSum	&	41.68	&	18.78	&38.00\\
%         BanditSum+entropy	&	41.71	&	18.87	&	38.04	\\
%         BanditSum+KL	&	\textbf{41.81*}	&	18.96*	&	\textbf{38.16*}	\\ \bottomrule
%     \end{tabular}
%     \caption{Results of our models and baselines. Scores which are significantly higher than 
%     BanditSum  with $p<0.001$ are marked with a *. We used bootstrap resampling method to test 
%     for significance.} % RNES is an auto-regressive model and our second proposed technique -- KL -- only works on non-auto-regressive model.}
%     %(details in Appendix \ref{significance})
%     \label{tab:results}
% \end{table}

We test for significance between the baseline models and our proposed techniques using the bootstrap method. This method was first recommended for testing significance in ROUGE scores by \citet{eva1_lin:2004:ACLsummarization}, and has subsequently been advocated as an appropriate measure in works such as \citet{dror-hitchhikers} and \citet{berg-kirkpatrick}.

\begin{figure}
    \centering
    \includegraphics[width=\linewidth]{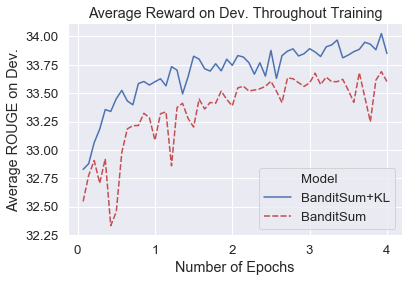}
    \caption{Training curves for BanditSum based models. Average ROUGE is the average of ROUGE-1, -2 and -L F1.}
    \label{fig:train_curves}
    \vspace{2mm}
\end{figure}

The simple entropy regularizer has a small but not significant improvement, and pretraining
has a similar improvement only for RNES. But the auxiliary ROUGE loss significantly ($p<0.001$) 
improves over BanditSum, obtaining an extra 0.15 ROUGE points on average. 
The last column 
reports the percentage of summary sentences which overlap with the lead. The auxiliary
loss leads to a 4.7\% absolute decrease in such selections compared 
to the base system, while also reaching a better ROUGE score. 
Figure \ref{fig:train_curves} shows that the reward (average ROUGE-1,-2,-L) for the
auxiliary loss model is consistently above the base.

We also examined the auxiliary loss model on documents where the summary is mostly comprised of 
lead sentences \Dearly{}, mostly sentences much later in the article \Dlate{}, and a dataset 
at the midway point, \Dmedian{}.
To create these sets, we rank test articles using the average index of its
summary sentences in the source document. The
100 test articles with lowest average index are \Dearly{}, the 100 with highest value 
are \Dlate{} and the 100 closest to the median are \Dmedian{}. 
%each with 100 documents with good extractive sentences appearing early/late in the document, respectively.  \Dearly{} should contain more lead bias. 
In Table~\ref{d_early}, we can see that the auxiliary loss model's improvements are even
more amplified on \Dmedian{} and \Dlate{}. 

On the other hand, our pretraining results are mixed. We hope to employ more controlled multi-tasking methods \citep{kiperwasser2018scheduled}
in the future to deal with the issue.

The second line in Table \ref{d_early} reports the oracle ROUGE scores of the best possible extractive summary.
% \footnote{The set of source sentences which obtains highest ROUGE score against the abstractive gold standard.}.
%AL2 how is the oracle computed (by similarity or is this taken from the dataset and was computed by someone else)
% MG: added explanation
While all systems are quite close to the oracle on \Dearly{}
they only reach half the performance on \Dlate{}. This gap indicates that our improvements only scratch the surface, but also that this problem is worthy and challenging to explore. 
% To get a sense of the content of these datasets, we analyzed the most frequent words occurring in their summaries (Table \ref{tab:word_dist}). 

It is worth noting that we have attempted to build a single model which can summarize both lead-biased articles and those
whose information is spread throughout. Our aim was to 
encourage the model to explore useful regions as a way of
learning better document semantics. But we hypothesize that
our models can be further improved by  learning to automatically predict when the lead paragraph 
suffices as a summary, and when
the model should look further in the document. 
%In preliminary experiments, 
%we have also found that if models were specifically 
%trained on \Dlate{} (selected using oracle 
%information as above) documents, a big boost in ROUGE scores
%can be obtained for these documents. 

\begin{table}
\centering
\small
\begin{tabular}{l|l|l|l}
\toprule
Model	&	$D_{\mathrm{early}}$	& $D_{\mathrm{med}}$ &	$D_{\mathrm{late}}$	\\ \hline
Lead-3	&	46.17	&	30.90	&	20.18	\\
Oracle	&	50.52	&	47.92	&	42.21	\\
% NeuSum	&	40.70	&	31.26	&	20.44	\\ \hline
RNES	&	41.76 &32.11 &20.62			\\ 
RNES+pretrain	& 41.66 & 32.38 & 20.64	\\ \hline
BanditSum	&	43.10	&	32.65	&	21.63	\\
BanditSum+entropy	&	41.96	&	32.59	&	22.12	\\
%BanditSum+pretraining &	43.13	&	32.70	&	21.68	\\
BanditSum+KL	&	42.63	&	33.05	&	21.96	\\ %\hline
%BanditSum_{early}   &  46.07 &  31.24   &   20.38   \\ 
%BanditSum_{late}    &  28.16  & 24.15  &  25.86 \\
%BanditSum+KL\_{early}	&	46.01	&	31.25	&	20.38	\\
%BanditSum+KL\_{late}	&	28.16	&	24.15	&	25.86	\\
\bottomrule
\end{tabular}
\caption{Average ROUGE-1, -2 and -L F1 scores on $D_{\mathrm{early}}$, and $D_{\mathrm{med}}$, $D_{\mathrm{late}}$. Each set contains 100 documents.}
\label{d_early}

% \small
% \begin{tabular}{l|l|l}
% \toprule
%  & 1st D$_{early}$ & 1st D$_{late}$ \\ \hline
% BanditSum                  & \textbf{50.5\% } & 47.4\%  \\ 
% BanditSum + pre-train      & 49.5\%  &\textbf{ 52.6 \% }\\ \bottomrule
% \end{tabular}
% \caption{Human evaluation on $D_{\mathrm{early}}^{20}$ and $D_{\mathrm{late}}^{20}$.}
% \label{table_human}
\end{table}
\vspace{2mm}

\section{Conclusion}
%\label{sec:discussion}
%\begin{itemize}
 %   \item test on other domain rather than news, when the lead-bias might not be that strong.
  %  \item compare to other model that don't use lead bias.
   % \item compare with models that use strong lead bias
    %\item test on other news corpora
    
%\end{itemize}

%AL2 update conclusions after obtaining final results. 

In this paper, we have presented the first approaches for learning a summarization system by countering the strong effect of summary-worthy lead sentences.
We demonstrate that recent summarization systems over-exploit the inherent lead bias present in news articles, to the detriment of their summarization capabilities. We explore two techniques aimed at learning to better balance positional cues with semantic ones.
While our auxiliary loss method achieves
significant improvement, we note that there is a large gap which better methods can
hope to bridge in the future.

One approach, building on ours, is to examine other ways to combine loss signals  \cite{maml}, and to encourage exploration \cite{sac}. 
We will also carry out deeper study of the properties 
of \Dearly{} and \Dlate{} type 
documents and use them to inform new solutions. 
On cursory analysis, the most frequent terms in \Dearly{} tend to be about UK politics, while  in \Dlate{} they are often related to British soccer.

\section*{Acknowledgments}
This work is supported by the Natural Sciences and Engineering Research Council of Canada, the Institute for Data Valorisation (IVADO), Compute Canada, and the CIFAR Canada AI Chair program. 
% The acknowledgments should go immediately before the references.  Do
% not number the acknowledgments section. Do not include this section
% when submitting your paper for review. \\

\bibliography{emnlp-ijcnlp-2019}
\bibliographystyle{acl_natbib}

\end{document}